\documentclass[11pt]{article}

\usepackage{acl}

\usepackage{times}
\usepackage{latexsym}
\usepackage[T1]{fontenc}
\usepackage[utf8]{inputenc}
\usepackage{microtype}
\usepackage{inconsolata}
\usepackage{graphicx}

\usepackage{amsmath}
\usepackage{amssymb}
\usepackage{amsthm}
\usepackage{booktabs}
\usepackage{multirow}
\usepackage{xcolor}

\newcommand{\Lorth}{\mathcal{L}_{\text{orth}}}
\newcommand{\mso}{\text{MSO}}

\title{Geometric Regularization in MoEs:\\The Disconnect Between Weights and Activations}

\author{Hyunjun Kim \\
  Korea Advanced Institute of Science and Technology (KAIST) \\
  \texttt{hyunjun1121@kaist.ac.kr}}

\begin{document}
\maketitle

\begin{abstract}
Mixture-of-Experts (MoE) models achieve efficiency through sparse activation, but the role of geometric regularization in expert specialization remains unclear.
We apply orthogonality loss to enforce expert diversity and find it \textit{fails} on multiple fronts: it does not reduce weight-space overlap (MSO actually \textbf{increases} by up to 114\%), \textbf{activation-space} overlap remains high ($\sim$0.6) regardless of regularization, and effects on performance are \textbf{inconsistent}---marginal improvement on WikiText-103 ($-0.9\%$), slight degradation on TinyStories ($+0.9\%$), and \textit{highly variable} results on PTB (std $>$ 1.0).
Our analysis across 7 regularization strengths reveals no significant correlation ($r = -0.293$, $p = 0.523$) between weight and activation orthogonality.
These findings demonstrate that weight-space regularization neither achieves its geometric goal nor reliably improves performance, making it unsuitable for MoE diversity.
\end{abstract}

\section{Introduction}
\label{sec:intro}

Mixture-of-Experts (MoE) models scale efficiently by activating only a subset of parameters per input \cite{shazeer2017outrageously,fedus2022switch}.
A common assumption is that expert representations should be \textit{orthogonal} to minimize interference \cite{chen2022towards,chen2023omoe}.
This intuition stems from linear algebra: orthogonal vectors are maximally distinguishable and their outputs do not interfere when combined.

\paragraph{Hypothesis.} Orthogonality regularization should improve expert diversity and reduce perplexity.

\paragraph{Finding.} It does not---and is unreliable. Across three datasets (TinyStories, WikiText-103, PTB), geometric regularization yields \textit{inconsistent} results: marginal improvement on WikiText-103 ($-0.9\%$), slight degradation on TinyStories ($+0.9\%$), and \textbf{high variance} on PTB (std $>$ 1.0).

\paragraph{Why?} We identify a \textbf{Weight-Activation Gap}: weight-space orthogonality (MSO $\approx 10^{-4}$) does not translate to activation-space orthogonality (MSO $\approx 0.6$). Across 7 regularization strengths, we find no significant correlation between weight and activation overlap ($r = -0.293$, $p = 0.523$), indicating that weight geometry and functional orthogonality are largely independent.

\paragraph{Contributions.}
\begin{enumerate}
    \item We show that orthogonality regularization \textit{fails} to reduce weight MSO---it actually increases it by up to 114\%---and yields \textit{inconsistent} effects on loss: marginal improvement on WikiText-103 ($-0.9\%$), slight degradation on TinyStories ($+0.9\%$), and high variance on PTB (std $>$ 1.0).
    \item We identify a weight-activation disconnect: activation overlap is $\sim$1000$\times$ higher than weight overlap, with no significant correlation ($r = -0.293$, $p = 0.523$, $n$=7).
    \item We demonstrate that weight-space regularization is an \textit{unreliable} optimization target---it neither achieves its geometric goal nor reliably improves performance.
\end{enumerate}

\section{Method}
\label{sec:method}

\paragraph{Orthogonality Loss.}
For expert weight matrices $\{W_i\}_{i=1}^N$, we define:
\begin{equation}
    \Lorth = \sum_{i < j} |\langle \tilde{W}_i, \tilde{W}_j \rangle|^2
    \label{eq:orth_loss}
\end{equation}
where $\tilde{W}_i = \text{vec}(W_i) / \|\text{vec}(W_i)\|$ is the normalized flattened weight vector. This loss encourages orthogonality among expert representations and is added to the language modeling objective with weight $\lambda$.

\paragraph{Mean Squared Overlap (MSO).}
We measure geometric diversity using:
\begin{equation}
    \mso = \frac{2}{N(N-1)} \sum_{i < j} |\langle \tilde{W}_i, \tilde{W}_j \rangle|^2
\end{equation}
Lower MSO indicates more orthogonal (diverse) experts. We compute MSO for both \textit{weights} and \textit{activations}.

\paragraph{Activation MSO.}
For co-activated experts producing outputs $\{h_i\}$, we compute:
\begin{equation}
    \mso_{\text{act}} = \mathbb{E}_{x}\left[\frac{2}{k(k-1)} \sum_{i < j \in \mathcal{S}(x)} \left(\frac{\langle h_i, h_j \rangle}{\|h_i\|\|h_j\|}\right)^2\right]
\end{equation}
where $\mathcal{S}(x)$ is the set of $k$ selected experts for input $x$. This measures functional similarity between expert outputs on actual inputs.

\section{Experiments}
\label{sec:exp}

\paragraph{Setup.}
We train NanoGPT-MoE ($\sim$130M parameters, 8 experts, 6 layers, top-2 routing) on TinyStories \citep{eldan2023tinystories} for 10K iterations with AdamW \citep{loshchilov2019adamw} (lr=$5 \times 10^{-4}$, $\beta_1$=0.9, $\beta_2$=0.95, weight decay=0.1). Each MoE layer contains 8 experts with hidden dimension 512 and intermediate dimension 2048. TinyStories experiments use 5 random seeds (42, 123, 456, 789, 1337).

\paragraph{Implementation Details.}
We regularize the \textbf{up-projection weights} ($W_{\text{up}} \in \mathbb{R}^{d_{\text{ffn}} \times d_{\text{model}}}$) of each expert. Each weight matrix is flattened and L2-normalized before computing pairwise inner products. The $\lambda$ sweep uses 7 values: $\{0, 0.001, 0.005, 0.01, 0.05, 0.1, 0.2\}$. MSO is computed \textbf{per layer} and \textbf{averaged} across all 6 MoE layers. Activation MSO is computed on the \textbf{post-gating} expert outputs for the top-2 selected experts, \textbf{unweighted} by gating scores. We do \textit{not} use auxiliary load balancing loss.

\subsection{Weight MSO Under Regularization}

Table~\ref{tab:weight_results} reveals a surprising finding: orthogonality regularization does \textit{not} reduce weight MSO---it \textbf{increases} it. Despite the loss explicitly penalizing expert overlap, the final weight MSO rises with $\lambda$, suggesting the regularization interferes with natural training dynamics rather than enforcing orthogonality.

\begin{table}[t]
\centering
\small
\begin{tabular}{lcc}
\toprule
\textbf{Method} & \textbf{Weight MSO} & \textbf{$\Delta$} \\
\midrule
Baseline & $5.43 \times 10^{-4}$ & --- \\
+ Orth ($\lambda$=0.001) & $7.52 \times 10^{-4}$ & $+39\%$ \\
+ Orth ($\lambda$=0.01) & $1.16 \times 10^{-3}$ & $+114\%$ \\
\bottomrule
\end{tabular}
\caption{Orthogonality regularization \textit{increases} weight MSO, contrary to its intended effect. Baseline weights are already near-orthogonal.}
\label{tab:weight_results}
\end{table}

\subsection{Perplexity Does Not Improve}

Despite the explicit regularization objective, perplexity improvements are \textbf{not statistically significant} (Table~\ref{tab:ppl_results}).

\begin{table}[t]
\centering
\small
\begin{tabular}{lccc}
\toprule
\textbf{Method} & \textbf{Val PPL} & \textbf{$\Delta$\%} & \textbf{p-value} \\
\midrule
Baseline & $5.94 \pm 0.08$ & --- & --- \\
+ Orth ($\lambda$=0.001) & $6.00 \pm 0.32$ & $+0.9\%$ & 0.727 \\
\bottomrule
\end{tabular}
\caption{Orthogonality regularization does not improve perplexity ($p = 0.727$, paired t-test, $n$=5 seeds). Variance increases from 0.08 to 0.32.}
\label{tab:ppl_results}
\end{table}

\paragraph{Why Are p-values High?}
The high p-value ($p = 0.727$) reflects both minimal effect size and increased variance. The baseline shows low std (0.08), while $\lambda$=0.001 increases variance to 0.32---a 4$\times$ increase that destabilizes training. The slight PPL increase (+0.9\%) is dwarfed by this variance, indicating the regularization adds noise without benefit.

\subsection{The Weight-Activation Gap}
\label{sec:gap}

Table~\ref{tab:gap} reveals the core finding: weight and activation geometry are fundamentally decoupled.

\begin{table}[t]
\centering
\small
\begin{tabular}{lccc}
\toprule
$\lambda$ & \textbf{Weight MSO} & \textbf{Act. MSO} & \textbf{Ratio} \\
\midrule
0 (baseline) & $5.43 \times 10^{-4}$ & $0.572$ & 1053$\times$ \\
0.001 & $7.52 \times 10^{-4}$ & $0.581$ & 773$\times$ \\
0.01 & $1.16 \times 10^{-3}$ & $0.577$ & 496$\times$ \\
0.1 & $2.04 \times 10^{-3}$ & $0.593$ & 290$\times$ \\
0.2 & $2.78 \times 10^{-3}$ & $0.564$ & 203$\times$ \\
\bottomrule
\end{tabular}
\caption{Activation MSO ($\sim$0.57) remains constant while weight MSO increases with $\lambda$. Pearson $r = -0.293$, $p = 0.523$ ($n$=7), indicating no significant correlation.}
\label{tab:gap}
\end{table}

\begin{figure}[t]
\centering
\includegraphics[width=\columnwidth]{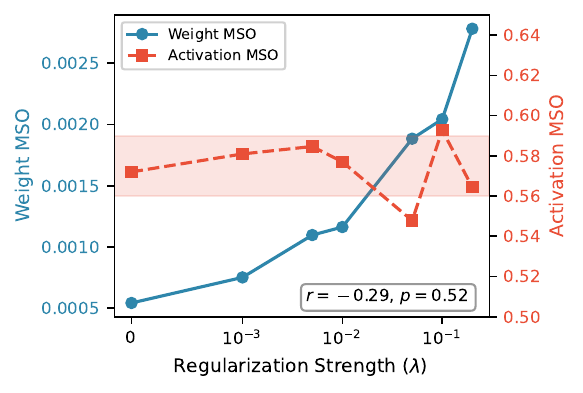}
\caption{Weight-Activation Gap. Weight MSO responds to regularization; activation MSO does not. No significant correlation ($r = -0.293$, $p = 0.523$).}
\label{fig:gap}
\end{figure}

\paragraph{Correlation Analysis.}
Figure~\ref{fig:gap} visualizes the disconnect: as $\lambda$ increases, weight MSO rises (regularization is applied), but activation MSO remains flat at $\sim$0.57. Across 7 regularization strengths, we find Pearson $r = -0.293$ ($p = 0.523$, 95\% CI: $[-0.857, 0.590]$)---\textit{not statistically significant}. This confirms that weight and activation geometry are independent.

\subsection{Cross-Dataset Validation}
\label{sec:cross_dataset}

To test whether our findings generalize beyond TinyStories, we evaluate orthogonality regularization on WikiText-103 \citep{merity2016pointer} and Penn Treebank \citep{marcus1993building}---two standard benchmarks with different characteristics (Table~\ref{tab:cross_dataset}).

\begin{table}[t]
\centering
\small
\begin{tabular}{lrccc}
\toprule
\textbf{Dataset} & \textbf{Tokens} & \textbf{Base} & \textbf{Orth} & \textbf{$\Delta$} \\
\midrule
WikiText-103 & 118M & $3.76_{\pm.02}$ & $3.73_{\pm.05}$ & $-0.9\%$ \\
TinyStories & 2.1M & $5.94_{\pm.08}$ & $6.00_{\pm.32}$ & $+0.9\%$ \\
PTB & 1.2M & $6.17_{\pm.95}$ & $5.74_{\pm1.11}$ & $-7.0\%^*$ \\
\bottomrule
\end{tabular}
\caption{Cross-dataset validation ($n$=3 seeds for WikiText-103/PTB, $n$=5 for TinyStories, $\lambda$=0.001). $^*$PTB shows high variance; effect direction unreliable.}
\label{tab:cross_dataset}
\end{table}

\paragraph{Dataset-Dependent Effects.}
Multi-seed experiments reveal mixed effects. On WikiText-103 (118M tokens), orthogonality regularization yields a small, consistent improvement ($-0.9\%$). However, on PTB (1.2M tokens), results are \textit{highly variable} across seeds (std $\sim$ 1.0), making conclusions unreliable. This high variance suggests that the effectiveness of geometric regularization may depend on dataset-seed interactions rather than dataset characteristics alone.

\paragraph{Interpretation.}
The high variance on PTB (std $\sim$1.0) compared to WikiText-103 (std $\sim$0.05) may reflect dataset-scale effects. Smaller datasets may exhibit more seed-dependent expert specialization patterns, leading to unstable outcomes. Regardless of direction, the \textit{inconsistency itself} undermines the reliability of weight-space regularization---a method that works only under specific seed-dataset combinations is impractical.

\section{Analysis: Why Does the Gap Exist?}
\label{sec:analysis}

\paragraph{The Role of Non-Linearities.}
Modern MoE experts use non-linear activation functions (SiLU/Swish) \citep{ramachandran2017swish} and LayerNorm \citep{ba2016layernorm}. Consider the expert computation:
\begin{equation}
    h_i = \text{LayerNorm}(\text{SiLU}(W_i \cdot x))
\end{equation}
Even if $W_i \perp W_j$, the non-linearities transform the geometry non-trivially. SiLU applies element-wise gating that depends on activation magnitudes, while LayerNorm re-centers and re-scales activations to unit variance. These operations can \textit{compress} the angular differences between expert outputs, increasing activation overlap regardless of weight geometry.

\paragraph{Input Distribution Effects.}
Weight orthogonality constrains \textit{static} parameters, but activation orthogonality depends on the \textit{input distribution}. If inputs $x$ project similarly onto different weight subspaces (e.g., due to low-rank structure in the input distribution), the resulting activations $W_i \cdot x$ and $W_j \cdot x$ will be correlated even when $W_i \perp W_j$. Natural language inputs may exhibit such structure due to semantic regularities.

\paragraph{Mathematical Intuition.}
Consider two weight matrices $W_1, W_2$ with Frobenius orthogonality $\langle W_1, W_2 \rangle_F = \text{tr}(W_1^T W_2) = 0$. This constraint only ensures the \textit{sum} of entries in $W_1^T W_2$ is zero---it does \textit{not} imply $W_1^T W_2 = 0$. For an input $x$, the activation inner product is $\langle z_1, z_2 \rangle = x^T W_1^T W_2 x$. Since $W_1^T W_2$ can have arbitrary non-zero structure (only its trace is constrained), this quadratic form is generally non-zero for typical inputs. Thus, Frobenius orthogonality of weights provides no guarantee of activation orthogonality.

\section{Related Work}
\label{sec:related}

\paragraph{MoE Scaling and Architecture.}
The modern MoE paradigm originates from \citet{shazeer2017outrageously}, who demonstrated trillion-parameter scaling via sparse gating. \citet{fedus2022switch} simplified this with top-1 routing in Switch Transformers, while GShard \citep{lepikhin2020gshard} enabled efficient distributed training. Recent work explores finer-grained expert decomposition: DeepSeekMoE \citep{dai2024deepseekmoe} uses 64 fine-grained experts per layer, and Mixtral \citep{jiang2024mixtral} achieves strong performance with 8 experts using top-2 routing. These architectures assume expert diversity emerges naturally through training.

\paragraph{Expert Specialization and Diversity.}
Several methods address expert diversity through routing improvements. X-MoE \citep{chi2022xmoe} uses hyperspherical routing with cosine-normalized gating to mitigate representation collapse. HyperRouter \citep{hyperrouter2024} dynamically generates router parameters via hypernetworks. S2MoE \citep{do2025s2moe} applies stochastic learning with Gaussian noise to prevent overlapping expert features. ReMoE \citep{remoe2025} proposes ReLU routing with L1 regularization for differentiable expert selection. SMoE-Dropout \citep{chen2023omoe} applies random routing to prevent expert collapse. CompeteSMoE \citep{pham2024competesmoe} uses competition-based routing to address representation collapse. Recent analysis by \citet{lo2025closer} finds that expert diversity increases with layer depth, yet concludes that the degree of expert specialization ``remains questionable.'' \citet{liu2023omoe} report that expert representations can exhibit up to 99\% similarity even in well-performing MoE models, proposing an orthogonal optimizer as a solution. Our results extend this skepticism: even when weight-space metrics suggest diversity, activation-space overlap persists. Critically, these methods primarily target \textit{routing} or \textit{optimization} diversity, whereas we analyze \textit{weight-space} geometric constraints directly.

\paragraph{Geometric Analysis in Deep Learning.}
Neural Collapse \citep{papyan2020prevalence,zhu2021geometric} shows that classifier representations converge to equiangular tight frames (ETFs) during terminal training. This geometric structure---where class representations are maximally separated---inspired our hypothesis that expert representations should similarly benefit from orthogonality. However, classifier heads are linear mappings from features to logits, whereas MoE experts are non-linear transformations with internal structure. Our negative result suggests the Neural Collapse analogy does not transfer: non-linearities break the geometry.

\section{Discussion}

Our results challenge the implicit assumption that weight-space orthogonality leads to functional diversity. Beyond being ineffective, geometric regularization is \textit{unreliable}---showing high variance (std $>$ 1.0) on smaller datasets like PTB.

\paragraph{Implications for MoE Design.}
\begin{itemize}
    \item \textbf{Weight-space regularization is unreliable.} Its effects are inconsistent: marginal improvement on WikiText-103 ($-0.9\%$), degradation on TinyStories ($+0.9\%$), and high variance on PTB. This unpredictability makes it unsuitable as a general strategy.
    \item \textbf{Activation-space regularization may be more appropriate.} Directly constraining $\mso_{\text{act}}$ during training could enforce functional diversity without the weight-activation disconnect.
    \item \textbf{Natural training already achieves low weight MSO.} The baseline weight MSO ($\sim 10^{-4}$) is already near-orthogonal, suggesting gradient descent implicitly regularizes expert geometry \citep{neyshabur2017implicit}.
    \item \textbf{Dataset scale matters.} Smaller datasets exhibit higher seed-dependent variance, making regularization effects unpredictable.
\end{itemize}

\section{Conclusion}

We investigate whether geometric regularization of MoE expert weights improves model performance and find that it does not---and is unreliable. Orthogonality loss \textit{fails} to reduce weight MSO (it increases by up to 114\%), and its effects on downstream performance are \textit{inconsistent}: marginal improvement on WikiText-103 ($-0.9\%$), slight degradation on TinyStories ($+0.9\%$), and high variance on PTB (std $>$ 1.0). We identify a fundamental disconnect between weight and activation orthogonality: activation MSO remains $\sim$1000$\times$ higher than weight MSO, with no significant correlation ($r = -0.293$, $p = 0.523$, $n$=7). This gap arises from non-linear transformations (SiLU, LayerNorm) and input distribution effects. Our analysis reveals that geometric regularization neither achieves its geometric goal nor reliably improves performance, making it unsuitable as a strategy for MoE diversity. Future work should explore activation-space regularization or alternative diversity metrics that directly target functional behavior.

\clearpage

\section*{Limitations}

\paragraph{Scale.} Our experiments use NanoGPT-MoE ($\sim$130M parameters). Whether the weight-activation gap persists at larger scales (1B+ parameters) or with different architectures (e.g., Mixtral, DeepSeek-MoE) remains an open question. Prior work reports router-level gains at larger scales \cite{wang2024auxfree}; our findings may be setup-specific.

\paragraph{Statistical Power.} TinyStories experiments use 5 random seeds with statistical testing. WikiText-103 experiments with 3 seeds show consistent improvement ($3.76 \pm 0.02$ baseline vs $3.73 \pm 0.05$ orth). PTB experiments exhibit \textit{high variance} across seeds ($6.17 \pm 0.95$ baseline vs $5.74 \pm 1.11$ orth), making conclusions about PTB unreliable. The single-seed result reporting degradation may not be representative.

\paragraph{Mechanism.} We identify the weight-activation gap but do not fully explain its cause. While we hypothesize that non-linearities and input structure play a role, a rigorous mathematical analysis of how SiLU and LayerNorm transform geometric relationships remains future work.

\paragraph{Baselines.} We do not compare against SMoE-Dropout \cite{chen2023omoe} or Loss-Free Balancing \cite{wang2024auxfree}, which may achieve different results through alternative mechanisms (random routing, auxiliary-loss-free bias updates).

\paragraph{Layer-wise Variation.} The weight-activation gap varies across layers: early layers (0--3) show \textit{larger} gaps ($\sim$3000--5600$\times$), while later layers (4--5) show \textit{smaller} gaps ($\sim$300--450$\times$). This pattern may arise from later layers having higher weight MSO ($\sim$10$\times$ higher than early layers), reducing the denominator of the ratio. A full analysis of why later layers develop less orthogonal weights remains future work.

\paragraph{Regularization Strength.} Our cross-dataset experiments use a fixed $\lambda=0.001$. The optimal regularization strength may vary across datasets; a comprehensive hyperparameter search could reveal conditions where geometric regularization is beneficial.

\paragraph{Future Directions.} Alternative approaches worth exploring include: (1) gradient-space orthogonality, enforcing that expert gradients point in different directions; (2) routing diversity losses that maximize expert selection entropy; (3) contrastive objectives that push apart expert outputs rather than weights; or (4) activation-space regularization that directly penalizes $\mso_{\text{act}}$ during training.

\bibliography{references}

\end{document}